\documentclass{article}
\usepackage{iclr2026_conference,times}
\usepackage{algpseudocode}

\usepackage{microtype}
\usepackage{graphicx}
\usepackage{subcaption}
\usepackage{booktabs} % for professional tables

\usepackage{minitoc}
\usepackage{placeins}
\usepackage[utf8]{inputenc} % allow utf-8 input
\usepackage[T1]{fontenc}    % use 8-bit T1 fonts
\usepackage{url}            % simple URL typesetting
\usepackage{booktabs}       % professional-quality tables
\usepackage{amsfonts}       % blackboard math symbols
\usepackage{nicefrac}       % compact symbols for 1/2, etc.
\usepackage{microtype}      % microtypography
\usepackage{xcolor}         % colors
\usepackage{pifont}
\usepackage[most]{tcolorbox}
\usepackage{subcaption} % For sub-figures
\usepackage{listings}
\usepackage{multirow}       % For multi-row cells
\usepackage{siunitx}
\definecolor{codegreen}{rgb}{0,0.6,0}
\definecolor{codegray}{rgb}{0.5,0.5,0.5}
\definecolor{codepurple}{rgb}{0.58,0,0.82}
\definecolor{backcolour}{rgb}{0.95,0.95,0.92}

\newcommand{\cmark}{\ding{51}}%
\definecolor{lightgray}{gray}{0.75}
\newcommand{\xmark}{\textcolor{lightgray}{\ding{55}}}% X mark symbol

\newcommand{\rot}[1]{\rotatebox{90}{#1}}
\newcolumntype{L}[1]{>{\raggedright\arraybackslash}p{#1}}

\usepackage[english]{babel}
\usepackage{fontawesome}
\usepackage{adjustbox} % Add this line to your preamble

\usepackage{booktabs,tabularx}

\definecolor{lighteconomist}{RGB}{252, 233, 237} % Adjust the RGB values to get the desired shade
\definecolor{economist}{RGB}{115,00,00} % Adjust the RGB values to get the desired shade
\definecolor{customgreen}{RGB}{116, 154, 114}
\definecolor{lightgreen}{RGB}{240, 246, 232}
\definecolor{greylight}{RGB}{242, 242, 242}
\definecolor{greydark}{RGB}{179, 179, 179}
\definecolor{ForestGreen}{RGB}{34, 139, 34}

\newcommand{\lightbulbicon}{%
  \begin{tikzpicture}[baseline=-0.5ex]
    \draw[fill=white, draw=insightteal, thick] (0,0) circle (1.5ex);
    \node[scale=0.8, color=insightteal] at (0,0) {\faLightbulbO~};
  \end{tikzpicture}%
}

\definecolor{insightteal}{RGB}{34, 139, 139}   % A sophisticated, medium-dark teal
\definecolor{insightback}{RGB}{240, 248, 248}   % A very light, complementary cyan-white

\usepackage{tikz} % needed for the ladder glyph
\usetikzlibrary{arrows.meta} % for Stealth arrow tips
\usetikzlibrary{decorations}
\usetikzlibrary{decorations.text}
\usetikzlibrary{decorations.pathreplacing}

\newcommand{\ladderrung}[2]{%
  \tikz[baseline=-0.65ex]{
    \ifnum#1>1
      \draw[greydark, line width=0.6pt] (0,0.70) -- (0,0.18);
    \fi
    \ifnum#1<#2
      \draw[greydark, line width=0.6pt] (0,-0.18) -- (0,-0.70);
    \fi
    \filldraw[white, draw=greydark, line width=0.6pt] (0,0) circle (1.55pt);
  }%
}
\definecolor{rowshade}{gray}{0.95}

\newtcolorbox{customblockquote}{
  colframe=insightteal,
  colback=insightback,
  boxrule=0pt,
  left=5pt,  % Set to 0pt so the background color touches the left 
  right=4pt,
  top=5pt,
  bottom=3pt,
  arc=0pt,
  breakable,
  before skip=1.2\baselineskip,
  after skip=0.7\baselineskip,
  left skip=0pt,
  right skip=0pt,
  enhanced jigsaw,
  frame hidden,
   overlay={
    \draw[insightteal, line width=2pt] 
      ([yshift=1pt]frame.north west) -- (frame.south west);
    \node[inner sep=0pt] at ([xshift=0pt, yshift=-1.3pt]frame.north west) {\lightbulbicon};
  },
  fontupper=\fontfamily{lmr}\selectfont,
  boxsep=1pt,
}
\usepackage{wrapfig}
\usepackage{rotating}
\usepackage{xcolor,colortbl}
\definecolor{verylightgray}{gray}{0.95}
\usetikzlibrary{backgrounds, calc, positioning, fit}

\newtcolorbox{greycustomblock}{
  colframe=greydark,        % Left line color
  colback=greylight,        % Background color
  boxrule=1pt,              % Line width
  left=2.5pt,               % Inner left margin
  right=3pt,                % Inner right margin
  top=5pt,                  % Inner top margin
  bottom=3pt,               % Inner bottom margin
  arc=0pt,                  % No rounded corners
  breakable,                % Allow breaking across pages
  before skip=0.2\baselineskip, % Vertical space before
  after skip=0.2\baselineskip,  % Vertical space after
  left skip=0pt,            % Left skip (adjust as needed)
  right skip=0pt,           % Right skip (adjust as needed)
  enhanced jigsaw,          % Enhanced jigsaw for precise control
  frame hidden,             % Hide the default frame
  overlay={                 % Custom overlay for left line
    \draw[greydark, line width=2pt]
      ([yshift=-1pt]frame.north west) -- ([yshift=1pt]frame.south west); % Adjusting the y coordinates to match exactly
  },
  fontupper=\selectfont, % Font size and style for the table
}
\usepackage{hyperref}

\usepackage{iclr2026_conference}

\iclrfinalcopy

\usepackage[utf8]{inputenc}
\usepackage[T1]{fontenc}
\usepackage{amsfonts}
\usepackage{amssymb}
\usepackage{amsmath}
\usepackage{nicefrac}
\usepackage{xcolor}
\usepackage{algorithm}
\usepackage{multirow}
\usepackage{siunitx}
\usepackage{array}

\author{%
  Paulius Rauba\\
  University of Cambridge\\
  \texttt{pr501@cam.ac.uk} 
  \And
  Claudio Fanconi\\
  University of Cambridge\\
  \texttt{caf83@cam.ac.uk} \\
  \And
  Mihaela van der Schaar\\
  University of Cambridge\\
  \texttt{mv472@cam.ac.uk} 
}

\title{Tiny Autoregressive Recursive Models}

\begin{document}

\maketitle

% Recursive Autoregressive Architectures

% Autoregressive Recursive Transformers
% What Matters for Deep Autoregressive Recursive Transformers?




\begin{abstract}
Tiny Recursive Models (TRMs) have recently demonstrated remarkable performance on ARC-AGI, showing that very small models can compete against large foundation models through a two-step refinement mechanism that updates an internal reasoning state $z$ and the predicted output $y$. Naturally, such refinement is of interest for any predictor; it is therefore natural to wonder whether the TRM mechanism could be effectively re-adopted in autoregressive models. However, TRMs cannot be simply compared to standard models because they lack causal predictive structures and contain persistent latent states that make it difficult to isolate specific performance gains. In this paper, we propose the Autoregressive TRM and evaluate it on small autoregressive tasks. To understand its efficacy, we propose a suite of models that gradually transform a standard Transformer to a Tiny Autoregressive Recursive Model in a controlled setting that fixes the block design, token stream, and next-token objective. Across compute-matched experiments on character-level algorithmic tasks, we surprisingly find that there are some two-level refinement baselines that show strong performance. Contrary to expectations, we find no reliable performance gains from the full Autoregressive TRM architecture. These results offer potential promise for two-step refinement mechanisms more broadly but caution against investing in the \textit{autoregressive} TRM-specific model as a fruitful research direction. 

%Autoregressive Transformers can allocate additional computation in non-equivalent ways: as \emph{untied depth} (adding more distinct layers), \emph{tied recurrent depth} (reusing a layer multiple times), or \emph{within-token refinement} (updating representations before predicting the next token). Recent hierarchical refinement models suggest that within-token refinement should improve generalization per unit compute under strict left-to-right prediction. However, prior comparisons often confound compute placement with changes to model structure, halting, readout, or the token sequence itself, obscuring causal attribution. We study compute placement in a controlled autoregressive setting that fixes the block design, token stream, and next-token objective, and measures compute directly in decoder block evaluations. We introduce a compute-placement ladder that varies one design choice at a time, including parameter sharing, step conditioning, halting strategy, and simple two-stream refinement. Across compute-matched experiments on character-level algorithmic tasks, untied depth and a flat two-stream recurrent baseline consistently outperform hierarchical within-token refinement, which provides no reliable gains and can substantially degrade performance. These results caution against treating token-internal refinement as a dependable route to improved autoregressive generalization at fixed compute.
\end{abstract}

% =========================
% (A) INTRO — “FOR” THE TOKEN-LEVEL REFINEMENT HYPOTHESIS (then stress-test it)
% =========================
\section{Introduction}

\begin{wrapfigure}{r}{0.58\textwidth}
    \centering
    \vspace{-15pt} % Adjust to align the top of the figure with the text
    \includegraphics[width=\linewidth, trim=2cm 0.2cm 0.7cm 0.4cm, clip]{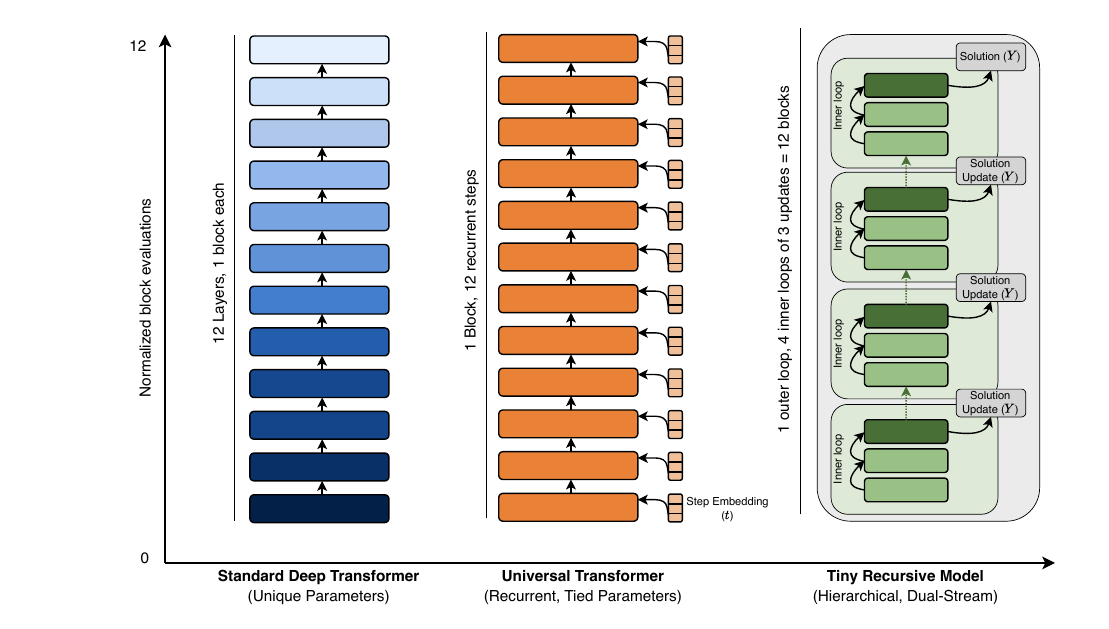}
    \caption{\textbf{Compute placement at fixed block-pass budget.} Three autoregressive decoders execute the same compute (12 decoder-block evaluations) but allocate it differently: \textbf{(left) Deep Transformer}: 12 untied layers, \textbf{(middle) Universal Transformer}: one shared block unrolled for 12 recurrent steps, with step embeddings, and \textbf{(right) Tiny Recursive Model}: hierarchical dual-stream refinement using multiple inner updates before each solution update. We investigate which allocation yields the best generalization per block evaluation.}
    \label{fig:fig1}
    \vspace{-20pt} % Adjust to reduce white space below the caption
\end{wrapfigure}
A common way to improve performance in autoregressive Transformers is to ``increase compute''. Yet, under a fixed decoder-block template, the same budget of decoder-block evaluations can be allocated in non-equivalent ways: (i) \emph{untied depth} (distinct layers), (ii) \emph{tied recurrent depth} (reusing a shared block across steps), or (iii) \emph{within-token refinement} (multiple internal updates before emitting next-token logits), as just a few examples. These choices match compute in block passes but differ in parameter sharing, state evolution, and how gradients supervise intermediate computation (see Fig. \ref{fig:fig1}).

Recent hierarchical refinement models motivate within-token refinement as latent multi-step computation, i.e. instead of emitting intermediate tokens, the model iteratively refines internal latents before producing logits~\cite{wang2025hierarchical, jolicoeur2025less}. The TRM model shows particularly strong results due to its strong performance in the ARC-AGI challenge suite relative to standard LLMs\footnote{Despite its impressive results, two key differences in the setup are that the TRM was trained as a supervised learning model directly on available data; and used extensive data augmentation for ARC-AGI}. Implicitly, this can be seen as a ``\textit{token-level reasoning}'' hypothesis, where reasoning here is used loosely as simply another word for computation. Therefore, if this token-level reasoning hypothesis holds, then under matched compute, allocating iteration \emph{within} each decoding step should improve generalization per block evaluation by enabling intra-step self-correction prior to committing a token. 

Can we directly test this hypothesis in other settings, such as the autoregressive setting? To understand why and when it works, we would require isolating the specific mechanism that the TRM adopts under matched compute. We therefore take \emph{compute placement} as the object of study and ask:
\emph{Given a fixed autoregressive decoder block template and a fixed next-token objective, how should iterative computation be allocated to maximize generalization per unit compute?}
We answer this in a controlled autoregressive family that fixes the token stream and next-token loss, enforces standard causal masking and KV-cache semantics, uses an identical decoder block, and excludes routing and token-stream modifications. Within this family, architectures correspond to different unrollings of the same block template.

\vspace{-2mm}
\paragraph{Findings.}
Across compute-matched experiments on character-level algorithmic tasks (addition, copying, reversing), untied depth and a \emph{flat} two-stream recurrent baseline yield the strongest generalization per block evaluation. The autoregressive TRM model, contrary to expectations, provides no consistent benefit. In fact, in most experiments, the performance degrades sharply. This suggests that under autoregressive setups, maintaining an inner refinement loop does not yield benefits given equivalent compute. However, we also recognize these are early results on ``tiny'' models in smaller data regimes; and these could, in principle, change, in not-so-tiny autoregressive recursive models.

\begin{customblockquote}
\textbf{Contributions.}
(i) We formalize \emph{compute placement} for autoregressive Transformers under a fixed block template and introduce a controlled ladder that isolates tying, step conditioning, halting/readout, and hierarchical refinement.
(ii) We derive an autoregressive projection of TRM-style hierarchical refinement that preserves causal masking and removes cross-call latent carry which allow for compute-matched comparisons without token-stream changes.
(iii) We empirically show that at matched block-pass budgets, untied depth and flat two-stream recurrence dominate, while token-internal hierarchical refinement is not a reliable route to improved autoregressive generalization.
\end{customblockquote}

\begin{figure}[t]
\centering
\resizebox{\textwidth}{!}{
\begin{tikzpicture}[
    >=Stealth,
    state/.style={circle, draw, minimum size=0.55cm, font=\tiny, inner sep=0pt, thick},
    label/.style={font=\scriptsize\sffamily\bfseries},
    sublabel/.style={font=\tiny\sffamily, text=black!55},
    arrow/.style={->, thick},
    dashedarrow/.style={->, dashed, thin},
    takeaway/.style={font=\tiny\sffamily, text=black!80, align=center, 
                     draw=black!15, rounded corners=2pt, fill=black!3, 
                     inner xsep=4pt, inner ysep=2pt}
]

% === Color Palette ===
\definecolor{streamH}{RGB}{70,130,180}    % Steel blue - hidden states
\definecolor{streamY}{RGB}{189,87,17}     % Warm brown - solution stream
\definecolor{streamZ}{RGB}{106,90,205}    % Slate purple - auxiliary stream
\definecolor{neutral}{RGB}{110,110,110}   % Gray

% ============ (a) Single-Stream Fixed ============
\begin{scope}[local bounding box=A]
\node[label] at (0, 2.0) {(a) No Recurrence};
\node[sublabel] at (0, 1.65) {Dense Transformer};

\node[state, fill=streamH!15, draw=streamH] (h0a) at (-1.4, 0.55) {$H^{0}$};
\node[state, fill=streamH!25, draw=streamH] (h1a) at (-0.45, 0.55) {$H^{1}$};
\node[state, fill=streamH!35, draw=streamH] (h2a) at (0.5, 0.55) {$H^{2}$};
\node[state, fill=streamH!50, draw=streamH] (hka) at (1.4, 0.55) {$H^{K}$};

\draw[arrow, streamH!80!black] (h0a) -- (h1a) node[midway, above, font=\tiny] {$f_\theta$};
\draw[arrow, streamH!80!black] (h1a) -- (h2a) node[midway, above, font=\tiny] {$f_\theta$};
\draw[arrow, streamH!80!black, densely dotted] (h2a) -- (hka);

\draw[decorate, decoration={brace, amplitude=3pt, mirror}, thick, neutral!70] 
    (-1.55, 0.05) -- (1.55, 0.05) node[midway, below=5pt, font=\tiny, text=neutral] {$K$ iterations};

% Takeaway
\node[takeaway] at (0, -0.65) {Readout: $H^{K}$};
\end{scope}

% ============ (b) Single-Stream Adaptive (UT) ============
\begin{scope}[shift={(4.5,0)}, local bounding box=B]
\node[label] at (0, 2.0) {(b) Single Stream};
\node[sublabel] at (0, 1.65) {Universal Transformer};

\node[state, fill=streamH!15, draw=streamH] (h0b) at (-1.4, 0.55) {$H^{0}$};
\node[state, fill=streamH!25, draw=streamH] (h1b) at (-0.45, 0.55) {$H^{1}$};
\node[state, fill=streamH!35, draw=streamH] (h2b) at (0.5, 0.55) {$H^{2}$};
\node[state, fill=streamH!50, draw=streamH] (hkb) at (1.4, 0.55) {$H^{K}$};

\draw[arrow, streamH!80!black] (h0b) -- (h1b);
\draw[arrow, streamH!80!black] (h1b) -- (h2b);
\draw[arrow, streamH!80!black, densely dotted] (h2b) -- (hkb);

% Weighted readout visualization
\node[state, fill=streamY!30, draw=streamY, minimum size=0.55cm] (out) at (0, -0.35) {};
\node[font=\tiny] at (0, -0.35) {$\Sigma$};

\draw[dashedarrow, neutral] (h0b) -- (out);
\draw[dashedarrow, neutral] (h1b) -- (out);
\draw[dashedarrow, neutral] (h2b) -- (out);
\draw[dashedarrow, neutral] (hkb) -- (out);

\node[font=\tiny, text=neutral] at (-1.55, 0.0) {$w_0$};
\node[font=\tiny, text=neutral] at (1.55, 0.0) {$w_K$};

% Takeaway
\node[takeaway] at (0, -1.0) {Readout: $\sum_k w_k H^{k}$};
\end{scope}

% ============ (c) Two-Stream ============
% ============ (c) Two-Stream ============
\begin{scope}[shift={(9.2,0)}, local bounding box=C]
\node[label] at (0, 2.0) {(c) Two-Stream};
\node[sublabel] at (0, 1.65) {Dual Universal Transformer};

% Y stream (solution) - top row
\node[state, fill=streamY!20, draw=streamY] (y0) at (-1.3, 0.9) {$Y^{0}$};
\node[state, fill=streamY!35, draw=streamY] (y1) at (0, 0.9) {$Y^{1}$};
\node[state, fill=streamY!50, draw=streamY] (y2) at (1.3, 0.9) {$Y^{K}$};

% Sigma below (define position early)
\node[state, fill=streamY!30, draw=streamY, minimum size=0.55cm] (outc) at (0, -0.75) {};
\node[font=\tiny] at (0, -0.75) {$\Sigma$};

% === DRAW ARROW BEFORE Z NODES so Z appears on top ===
\draw[dashedarrow, neutral] (y1.south) -- (outc.north);

% Z stream (auxiliary) - drawn AFTER arrow so Z1 is on top
\node[state, fill=streamZ!20, draw=streamZ] (z0) at (-1.3, 0.0) {$Z^{0}$};
\node[state, fill=streamZ!35, draw=streamZ] (z1) at (0, 0.0) {$Z^{1}$};
\node[state, fill=streamZ!50, draw=streamZ] (z2) at (1.3, 0.0) {$Z^{K}$};

% Y-stream transitions (horizontal)
\draw[arrow, streamY!80!black] (y0) -- (y1);
\draw[arrow, streamY!80!black, densely dotted] (y1) -- (y2);

% Z-stream transitions (horizontal)
\draw[arrow, streamZ!80!black] (z0) -- (z1);
\draw[arrow, streamZ!80!black, densely dotted] (z1) -- (z2);

% Cross-conditioning: Z -> Y
\draw[dashedarrow, streamZ!70!black] (z0.north east) -- (y1.south west);
\draw[dashedarrow, streamZ!70!black] (z1.north east) -- (y2.south west);

% Cross-conditioning: Y -> Z
\draw[dashedarrow, streamY!70!black] (y0.south east) -- (z1.north west);
\draw[dashedarrow, streamY!70!black] (y1.south east) -- (z2.north west);

% Takeaway
\node[takeaway] at (0, -1.35) {Readout: $\sum_k w_k Y^{k}$};
\end{scope}

% ============ (d) Nested Two-Stream ============
\begin{scope}[shift={(13.6,0)}, local bounding box=D]
\node[label] at (0, 2.0) {(d) Two Stream with Q-Halt};
\node[sublabel] at (0, 1.65) {Autoregressive TRM};

% Y stream (outer)
\node[state, fill=streamY!20, draw=streamY] (y0d) at (-1.3, 1.0) {$Y^{0}$};
\node[state, fill=streamY!50, draw=streamY] (y1d) at (1.3, 1.0) {$Y^{1}$};

% Inner Z refinements
\node[state, fill=streamZ!15, draw=streamZ, minimum size=0.42cm] (z00) at (-0.7, 0.2) {\tiny$Z^{0}$};
\node[state, fill=streamZ!30, draw=streamZ, minimum size=0.42cm] (z01) at (0.0, 0.2) {\tiny$Z^{1}$};
\node[state, fill=streamZ!45, draw=streamZ, minimum size=0.42cm] (z0L) at (0.7, 0.2) {\tiny$Z^{L}$};

\draw[arrow, streamZ!70!black, thin] (z00) -- (z01);
\draw[arrow, streamZ!70!black, thin, densely dotted] (z01) -- (z0L);

\draw[arrow, streamY!80!black, thick] (y0d) to[out=-20, in=160] (y1d);
\draw[dashedarrow, neutral] (z0L) -- (y1d);

% Annotations
\node[font=\tiny, text=streamZ!80!black] at (0, -0.2) {$L$ inner steps};
\node[font=\tiny, text=streamY!80!black, anchor=west] at (1.55, 1.0) {$\times H$};

% Takeaway
\node[takeaway] at (0, -0.65) {Readout: $Y^{(H)}$};
\end{scope}

% Dotted separators
\draw[dotted, thick, black!12] (2.65, -1.55) -- (2.65, 2.35);
\draw[dotted, thick, black!12] (6.85, -1.55) -- (6.85, 2.35);
\draw[dotted, thick, black!12] (11.4, -1.55) -- (11.4, 2.35);

% ============ LEGEND ============
% Horizontal separator line
\draw[black!20, thick] (-1.8, -1.60) -- (14.8, -1.60);

\node[anchor=west, font=\tiny\sffamily\bfseries, text=black!70] at (-1.6, -1.90) {States:};

\node[state, fill=streamH!30, draw=streamH, minimum size=0.32cm] at (-0.2, -1.90) {};
\node[anchor=west, font=\tiny\sffamily, text=black!70] at (0.0, -1.90) {Hidden $H$};

\node[state, fill=streamY!35, draw=streamY, minimum size=0.32cm] at (2.0, -1.90) {};
\node[anchor=west, font=\tiny\sffamily, text=black!70] at (2.2, -1.90) {Solution $Y$};

\node[state, fill=streamZ!35, draw=streamZ, minimum size=0.32cm] at (4.2, -1.90) {};
\node[anchor=west, font=\tiny\sffamily, text=black!70] at (4.4, -1.90) {Auxiliary $Z$};

\node[state, fill=neutral!25, draw=neutral, minimum size=0.32cm] at (6.4, -1.90) {};
\node[anchor=west, font=\tiny\sffamily, text=black!70] at (6.6, -1.90) {Input $X$};

\node[anchor=west, font=\tiny\sffamily\bfseries, text=black!70] at (8.2, -1.90) {Edges:};

\draw[arrow, streamH!80!black] (9.4, -1.90) -- (10.0, -1.90);
\node[anchor=west, font=\tiny\sffamily, text=black!70] at (10.1, -1.90) {Transition $f_\theta$};

\draw[dashedarrow, neutral] (12.4, -1.90) -- (13.0, -1.90);
\node[anchor=west, font=\tiny\sffamily, text=black!70] at (13.1, -1.90) {Conditioning};

\end{tikzpicture}
}
\caption{\textbf{Compute placement architectures.} 
(a)~Single hidden-state stream with final-iterate readout. 
(b)~Adaptive halting with weighted readout across all iterates. 
(c)~Two-stream factorization: solution $Y$ and auxiliary $Z$ with cross-conditioning. 
(d)~Nested hierarchy: $L$ inner refinements of $Z$ per outer update of $Y$, repeated $H$ times.}
\label{fig:architectures}
\end{figure}
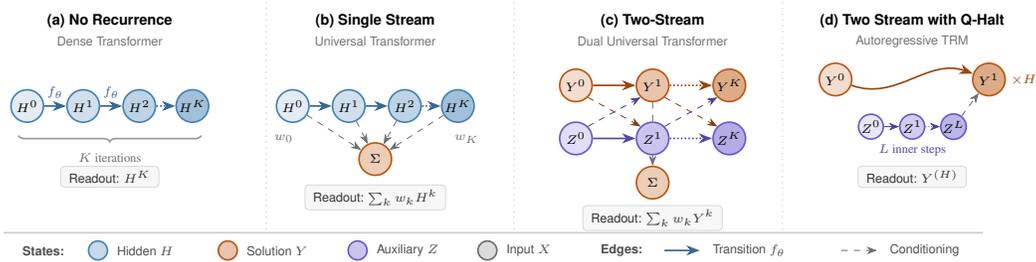

\vspace{-2mm}
\section{Background}

\paragraph{Preliminaries (autoregressive setting).}
Let $\mathcal{V}$ be a finite vocabulary and $\mathbf{x}=(x_1,\dots,x_T)\in\mathcal{V}^T$ a length-$T$ sequence. We model
\[
p_\theta(\mathbf{x})=\prod_{t=1}^{T} p_\theta(x_t \mid \mathbf{x}_{<t}),
\qquad \mathbf{x}_{<t}=(x_1,\dots,x_{t-1}),
\]
and train by maximum likelihood,
\[
\mathcal{L}(\theta)= -\sum_{t=1}^{T}\log p_\theta(x_t \mid \mathbf{x}_{<t}).
\]
At depth $\ell$, hidden states are $H^{(\ell)}\in\mathbb{R}^{T\times d}$ with rows $h_t^{(\ell)}\in\mathbb{R}^d$, and
$H^{(0)}=\mathrm{Embed}(\mathbf{x})+\mathrm{Pos}(\mathbf{x})$ (optionally plus other additive signals). We use causal language modelling, enforcing causality with an attention mask that restricts each position to attend only to earlier tokens.

\subsection{Transformers from a compositional point of view}
Standard Transformers~\citep{vaswani2017attention} implement $\mathbf{x}_{<t}\mapsto p_\theta(x_t\mid \mathbf{x}_{<t})$ by stacking $L$ residual blocks. Each block $\ell\in\{0,\dots,L-1\}$ applies $f_{\theta_\ell}:\mathbb{R}^{T\times d}\to\mathbb{R}^{T\times d}$ (attention--MLP), giving
\[
H^{(\ell+1)} = f_{\theta_\ell}\!\left(H^{(\ell)}\right),
\qquad
H^{(L)} = f_{\theta_{L-1}}\circ\cdots\circ f_{\theta_0}\!\left(H^{(0)}\right).
\]
Equivalently, a Transformer repeatedly applies a block template, composing $L$ distinct transformations in sequence. This fixed-depth, acyclic graph is easy to optimize and parallelize, but hard-codes a uniform compute budget: each token receives exactly $L$ block applications. Any additional iterative computation must therefore be realized indirectly (e.g., via extra tokens, a common technique for test-time scaling) or by modifying the architecture to allocate compute differently.

\subsection{Universal Transformers}
\label{subsec:UTs}
Universal Transformers (UTs)~\citep{dehghani2018universal} replace untied depth with recurrent depth by repeatedly applying a single block $f_\theta$:
\[
H^{(k+1)} = f_\theta(H^{(k)}), \qquad k=0,1,\dots,K-1,
\]
with $H^{(0)}$ as above. For fixed $K$, this is a depth-$K$ Transformer with full parameter tying, increasing computational steps without increasing parameters and interpreting recurrence as iterative refinement of a shared latent state. Because tying makes refinement steps ambiguous, UTs inject a step encoding so $f_\theta$ can condition on $k$.

UTs are typically combined with Adaptive Computation Time (ACT)~\citep{graves2016adaptive}, which adds a halting mechanism and forms outputs via a weighted accumulation of intermediate states with a ponder-style penalty. In principle this enables non-uniform compute across tokens. In practice, refinement is still coupled to a global recurrent mechanism (tokens share the same update sequence until halting), and predictions come from an aggregate over iterates rather than a final refined state. Thus, while UTs show recurrent depth can be effective, they do not resolve our finer question: under a fixed autoregressive decoder block, where does iteration improve generalization per unit compute?

\subsection{Tiny Recursive Models: An extension of UT}
\label{subsec:trms}
Tiny Recursive Models (TRMs)~\citep{jolicoeur2025less} have drawn attention because a 7M-parameter TRM reportedly generalizes well on ARC-AGI (e.g., 45\% on ARC-AGI-1 and 8\% on ARC-AGI-2 with heavy augmentation and test-time compute), suggesting recursive depth can support structured reasoning with small parameter budgets.

TRMs can be seen as extending UTs by keeping a tied refinement operator but changing \emph{what} is refined and \emph{how} it is read out. They decompose state into a fixed input stream $X=\mathrm{Embed}(\mathbf{x})+\mathrm{Pos}(\mathbf{x})$, a solution stream $Y$, and an auxiliary reasoning stream $Z$. With a shared block $f_\theta$, $Z$ is refined while conditioning on $X$ and the current solution, and $Y$ is updated through the refined $Z$:
\[
\begin{aligned}
Z^{(k,i+1)} &= f_\theta\!\big(X + Y^{(k)} + Z^{(k,i)}\big), \qquad i=0,\dots,n-1, \\
Y^{(k+1)} &= f_\theta\!\big(Y^{(k)} + Z^{(k,n)}\big).
\end{aligned}
\]
This induces a hierarchical schedule: multiple inner refinements of $Z$ precede each update of $Y$ (repeated over an outer loop). TRM halting also differs from ACT~\citep{graves2016adaptive}: instead of a convex combination of iterates with a ponder penalty, TRMs use a binary halt head on the solution stream and output the terminal iterate $Y^{(K)}$.

We use TRMs here for a narrower purpose: they provide a clean way to place iterative computation while keeping the block and objective fixed. However, in the original implementation, TRMs are introduced as supervised learning solvers, whereas autoregressive decoding must emit next-token distributions at every step. Emitting explicit ``thinking'' tokens would change the token stream and thus the learning problem, so instead we adapt TRM refinement to operate \emph{within} a masked decoder step: for a fixed prefix, we iteratively refine the hidden state producing next-token logits without extra tokens or routing (Sec.~\ref{sec:autoregressive_refinement}). Figure~\ref{fig:architectures} summarizes the architectures.

%Since causal decoding requires a single refinement depth per forward pass, we implement halting at the sequence level (shared across positions) via a binary head on the solution stream, so that increased test-time compute corresponds only to additional applications of the same tied operator. This makes TRMs a convenient baseline for our study: they instantiate ``token-internal'' iterative computation beyond recurrent depth, while preserving the strictly controlled family of autoregressive models that we aim to compare.

\section{Autoregressive token-level refinement}
\label{sec:autoregressive_refinement}

We return to the central question: \emph{under matched compute, how should iterative computation be allocated within an autoregressive decoder?} Existing approaches increase untied depth, reuse a block as recurrent depth~\citep{dehghani2018universal}, or add within-step refinement intended to resemble ``token-level reasoning''~\citep{wang2025hierarchical,chen2025inner}. However, these proposals are rarely directly comparable: they often change the token stream (e.g., scratch/pause tokens)~\citep{wei2022chain,nye2021show,zelikman2022star}, add conditional capacity via routing~\citep{shazeer2017outrageously,lepikhin2020gshard,fedus2022switch}, or alter the inference interface (e.g., solver-style refinement)~\citep{bai2019deep}, confounding where supervision acts. Methods that ``think'' by emitting extra tokens may help~\citep{wei2022chain,nye2021show,zelikman2022star}, but they also change the modelling problem by changing the generated sequence. Here we fix the token stream and next-token objective, and ask whether extra computation can be organized \emph{internally} under strict autoregressive semantics.

Within this controlled scope, TRMs are of interest because they represent an extreme compute placement: refining a structured latent state multiple times \emph{within} a single decoding step. If token-internal refinement yields gains in our controlled autoregressive family~\citep{wang2025hierarchical,chen2025inner}, the autoregressive TRM model ought to show empirical performance gains. The mismatch is that TRMs are proposed as supervised solvers with bidirectional attention and cross-call state carry. This section adapts TRM-style refinement to autoregressive decoding and uses it to study compute placement in autoregressive Transformers.

\subsection{Compute placement and why TRM-style is the right stress test}
\label{subsec:compute_placement_view}

Autoregressive compute can be increased in non-equivalent ways: as untied depth (distinct blocks), tied recurrent depth (reapplying a shared block)~\citep{dehghani2018universal}, or within-step refinement (multiple internal updates before logits)~\citep{wang2025hierarchical,jolicoeur2025less}. These mechanisms can match block evaluations per token, but they differ in the iterated state and in how predictions are read out.

This motivates a pragmatic taxonomy used throughout the paper. \emph{Single-stream} variants differ by weight tying, step disambiguation, and whether intermediate iterates contribute via a weighted readout~\citep{dehghani2018universal,graves2016adaptive,banino2021pondernet}. \emph{Multi-stream} variants~\citep{wang2025hierarchical,jolicoeur2025less} decompose the latent state into asymmetric components (e.g., solution and auxiliary streams). This is a form of nested scheduling in which the auxiliary stream undergoes multiple inner refinements before each solution update\footnote{This contrasts with alternative nesting mechanisms that focus on parameter-level nesting \citep{rauba2026deep, behrouz2025nested}}. A final axis is the readout: aggregate over iterates or use a terminal iterate~\citep{graves2016adaptive,banino2021pondernet,jolicoeur2025less}.

\subsection{Two challenges to overcome}
\label{subsec:two_hurdles}

TRMs are not directly compatible with strict autoregressive decoding for two reasons.  $\blacktriangleright$ \textbf{Challenge 1 (causality).} Solver-style TRMs use bidirectional attention, so an update at position $i$ can depend on future positions $j>i$. We enforce strict autoregressive semantics by applying a causal mask. $\blacktriangleright$ \textbf{Challenge 2 (cross-call carry).} TRMs maintain a persistent latent carry that is reset only upon halting. This can leak information across prefixes: computation performed under earlier contexts can influence later logits beyond the explicit conditioning set $\mathbf{x}_{<t}$. We remove this by re-initializing the TRM latent streams on every forward pass, so logits at step $t$ depend only on $\mathbf{x}_{<t}$ within that call.

These modifications isolate token-internal refinement within a fixed compute budget. For a given block-pass budget, the nested schedule simply partitions block evaluations across inner $Z$ refinements and outer $Y$ updates. Consequently, any difference from single-stream recurrence cannot be attributed to future leakage, cross-call state, routing, or token-stream changes~\citep{shazeer2017outrageously,lepikhin2020gshard,fedus2022switch,wei2022chain,nye2021show,zelikman2022star}. In our compute-placement ladder (Sec.~\ref{sec:compute_placement}), this autoregressive TRM-style projection is the extreme token-internal rung; if it fails to improve generalization at equal compute, it provides a strong negative test for within-step refinement in this setting.
 
\vspace{-2mm}
\section{Mechanistic decomposition of iterative computation}
\label{sec:compute_placement}

\begin{wraptable}{r}{0.55\textwidth}
\centering
\vspace{-12pt} % Adjust to align table top with surrounding text
\scriptsize
\setlength{\tabcolsep}{4.5pt} % Reduced slightly for better fit
\renewcommand{\arraystretch}{1.2}
\begin{tabular}{lcccccc}
\toprule
\textbf{Model} & \rot{Tie} & \rot{Step} & \rot{ACT} & \rot{2S} & \rot{Nest} & \rot{Q-halt} \\
\midrule
\midrule
Dense Transformer              & \xmark & \xmark & \xmark & \xmark & \xmark & \xmark \\
Iterative Transformer          & \cmark & \xmark & \xmark & \xmark & \xmark & \xmark \\
Iterative Step Transformer     & \cmark & \cmark & \xmark & \xmark & \xmark & \xmark \\
Universal Transformer          & \cmark & \cmark & \cmark & \xmark & \xmark & \xmark \\
Dual UT                        & \cmark & \cmark & \cmark & \cmark & \xmark & \xmark \\
Dual Nested UT                 & \cmark & \cmark & \cmark & \cmark & \cmark & \xmark \\
Autoregressive TRM    & \cmark & \cmark & \cmark / \xmark & \cmark & \cmark & \cmark / \xmark  \\
\bottomrule
\end{tabular}
\vspace{3pt}
\caption{\footnotesize\textbf{Progressive deltas (single-column ladder).}
Each row adds exactly one mechanism relative to the previous row. Tie=weight tying; Step=step embeddings; ACT=Graves ACT-style halting; 2S=two-stream ($y,z$); Nest=hierarchical $H \times L$ loop; Q-halt=TRM-style binary halting. \textbf{Note:} We have used terms that best describe the underlying model architecture; for instance, Autoregressive TRM is best described for what it is doing, while the non-autoregressive version is called TRM.}
\label{tab:deltas_ladder_small}
\vspace{-10pt} % Reduces white space below the caption
\end{wraptable}

We study \emph{compute placement} in autoregressive Transformers: modifying how computation is allocated while holding fixed tokenization, the next-token objective, causal attention/KV-cache semantics, and the decoder-block template (attention--MLP with residual/norm), with no routing and no token-stream modifications. This controlled setting isolates where iteration matters.

We define three criteria that we think a study with high internal validity should therefore satisfy.
$\blacktriangleright$ \textbf{Controlled family:} fixed token stream, objective, masking, and block structure, so variants differ only in iteration placement;
$\blacktriangleright$ \textbf{Axis isolation:} vary one mechanism at a time (e.g., tying, halting, readout), enabling causal attribution;
$\blacktriangleright$ \textbf{Compute normalization:} match compute by counting block evaluations rather than parameters.

Many existing approaches vary multiple axes at once: standard Transformers use untied depth, single-stream state, and final-state readout; UTs add weight tying and step embeddings; ACT and PonderNet add adaptive halting with weighted accumulation; TRMs add state decomposition, hierarchical iteration, and binary halting with terminal readout; and Deep Equilibrium Models replace explicit unrolling with implicit fixed-point solvers, changing both iteration and training dynamics. These bundled changes make mechanistic attribution difficult.

\textbf{How to interpret Table~\ref{tab:deltas_ladder_small}.}
The table defines the controlled \emph{compute-placement ladder}: each row adds exactly one mechanism while holding fixed the token stream and next-token objective, causal masking/KV-cache semantics, and the decoder-block template $f_\theta$. Columns indicate:
$\blacktriangleright$\textsc{Tie}: reuse the same parameters across the $C$ block passes (tied recurrence) vs.\ distinct $\theta_\ell$ (untied depth);
$\blacktriangleright$\textsc{Step}: inject a step signal $e_k$ so a tied block can condition on iteration index;
$\blacktriangleright$\textsc{ACT}: Graves-style adaptive computation time, i.e.\ halting with weighted readout over iterates $H^{(k)}$ (run in a compute-matched regime when required);
$\blacktriangleright$\textsc{2S}: replace single-stream $H$ with two-stream $(Y,Z)$ conditioned on the fixed input stream $X=\mathrm{Embed}(\mathbf{x})+\mathrm{Pos}(\mathbf{x})$;
$\blacktriangleright$\textsc{Nest}: refine $Z$ for $L$ inner steps before each $Y$ update (repeated $H$ times), moving compute \emph{within} a decoding step without changing the token stream;
$\blacktriangleright$\textsc{Q-halt}: replace ACT-style accumulation with TRM-style binary halting and terminal-iterate readout (logits from the final solution iterate).
Under compute normalization (Sec.~\ref{sec:exp_framework}), all rows execute the same number of block evaluations per forward pass, so differences can be attributed to the single added mechanism.

\section{Evaluation}
\label{sec:evaluation}

\textbf{Evaluation methodology}. We evaluate a controlled ladder of seven autoregressive architectures that differ only in \emph{compute placement}, while holding fixed the token stream, next-token objective, causal masking/KV-cache semantics, and an identical decoder-block template (Pre-Layer-Normalization causal self-attention + GELU MLP with residuals), with no routing and no token-stream modifications. Compute is normalized by directly matching the number of decoder-block evaluations per forward pass (“block passes”) to a fixed budget $C$, configuring each model’s unrolling so that Dense uses $C$ untied layers, tied/step-aware variants reuse a shared block for $C$ steps (optionally with step embeddings), ACT-style UT variants are run in a full-compute mode (no early halting) to keep realized block passes fixed, and two-stream / nested variants subdivide the $C$ passes across reasoning/solution updates (and inner refinements when nested). All models are trained under identical optimization and data conditions (procedurally generated character-level algorithmic sequences; next-token cross-entropy with loss on outputs only), and are evaluated on Copy, Reverse, and Addition with out-of-distribution length scaling using greedy decoding. Our evaluation metrics include sequence exact match, character accuracy, and position-wise accuracy by output quartiles to localize where failures accumulate. We do not compute the sensitivity of the autoregressive samplers to input perturbations \citep{rauba2025statistical} because we only care about exactly predicting the single ground-truth answer given the input sequence.

%\vspace{2pt}
\subsection{Performance under equal compute}

\begin{figure*}[t]
    \centering
    \includegraphics[width=0.8\textwidth]{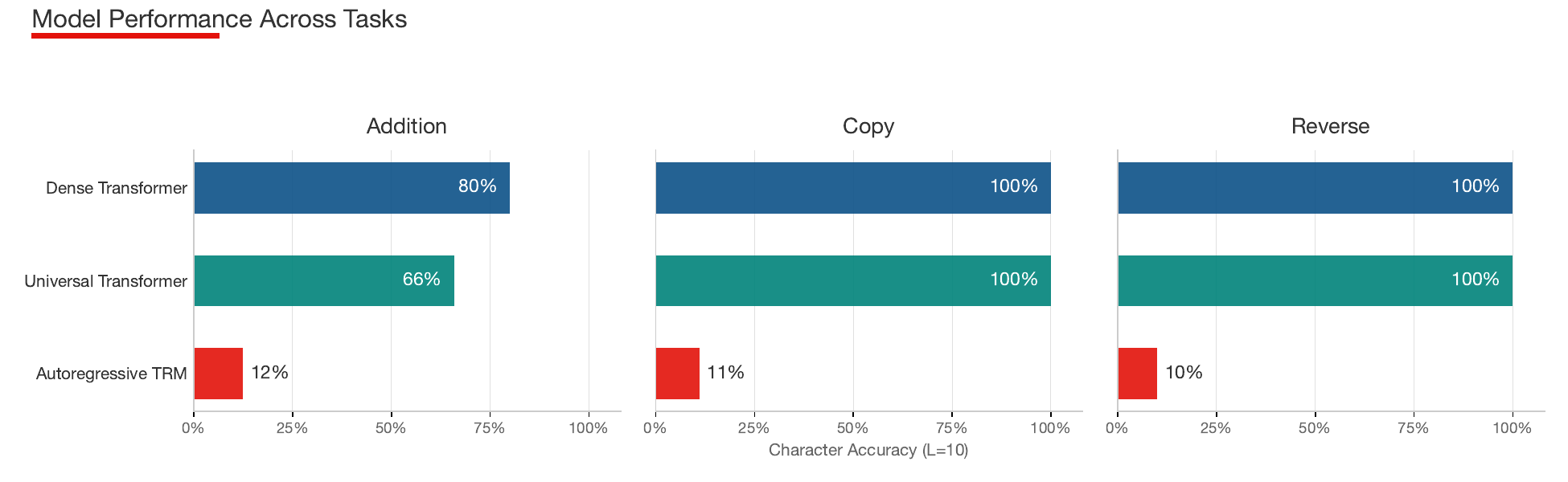}
    \caption{\textbf{Model performance across tasks.} Character accuracy (length $10$) on Addition, Copy, and Reverse for three architectures: Dense Transformer, Universal Transformer (UT), and autoregressive TRM. Copy and Reverse are solved by the Dense Transformer and UT (100\% accuracy), while Addition remains more difficult and separates these two models (80\% vs.\ 66\%). The autoregressive TRM performs poorly on all three tasks, reaching only 12\%, 11\%, and 10\% accuracy, respectively.}
    
    \label{fig:task-summary}
\end{figure*}

\begin{greycustomblock}
\textbf{RQ1}. Under matched block-pass compute, which compute placements yield the strongest performance, and which tasks most strongly differentiate these choices?
\end{greycustomblock}

\paragraph{Result}
Figure~\ref{fig:task-summary} compares three primary architectures on Addition, Copy, and Reverse. The Dense Transformer and UT both achieve perfect performance on Copy and Reverse, indicating that these two tasks are easy for these models at the tested sequence length. Addition is more demanding: the Dense Transformer reaches 80\% character accuracy, while the UT reaches 66\%. By contrast, the autoregressive TRM performs poorly on every task, with accuracy near 10\% throughout, which is close to chance-level behavior at the character level\footnote{The associated github repo can be found at https://github.com/pauliusrauba/autoregressive-TRM}.

\paragraph{Interpretation.}
The main pattern is that task demands interact strongly with architectural constraints. Copy and Reverse mainly test whether the model can preserve or reorder information already present in the input. Under the present training and evaluation setup, both the Dense Transformer and UT have sufficient capacity to do this reliably. Addition has a different structure: correct output requires intermediate state updates that remain consistent across positions, because local mistakes can propagate through the sequence via carry-like dependencies. This makes Addition a more sensitive test of whether the architecture can sustain a stable internal computation over multiple steps. The gap between the Dense Transformer and UT therefore shows that these two models, although both strong on Copy and Reverse, differ in how effectively they support that computation. The autoregressive TRM fails even on Copy and Reverse, which suggests that its difficulty is more basic: in this setting, it does not learn a robust input-output mapping for any of the three tasks.

\begin{customblockquote}
\textbf{Takeaway.} Copy and Reverse distinguish the autoregressive TRM from the other two models, while Addition further separates the Dense Transformer from the UT. The Dense Transformer performs best overall, the UT is strong but weaker on Addition, and the autoregressive TRM does not generalize successfully on any task shown.
\end{customblockquote}

\subsection{Error concentration and learning bottlenecks}

\begin{greycustomblock}
\textbf{RQ2}. Where do errors concentrate in autoregressive decoding, and when does training succeed (or fail) to resolve the final dependency under different compute placements?
\end{greycustomblock}

\begin{figure*}[h]
    \centering
            \includegraphics[width=0.95\textwidth]{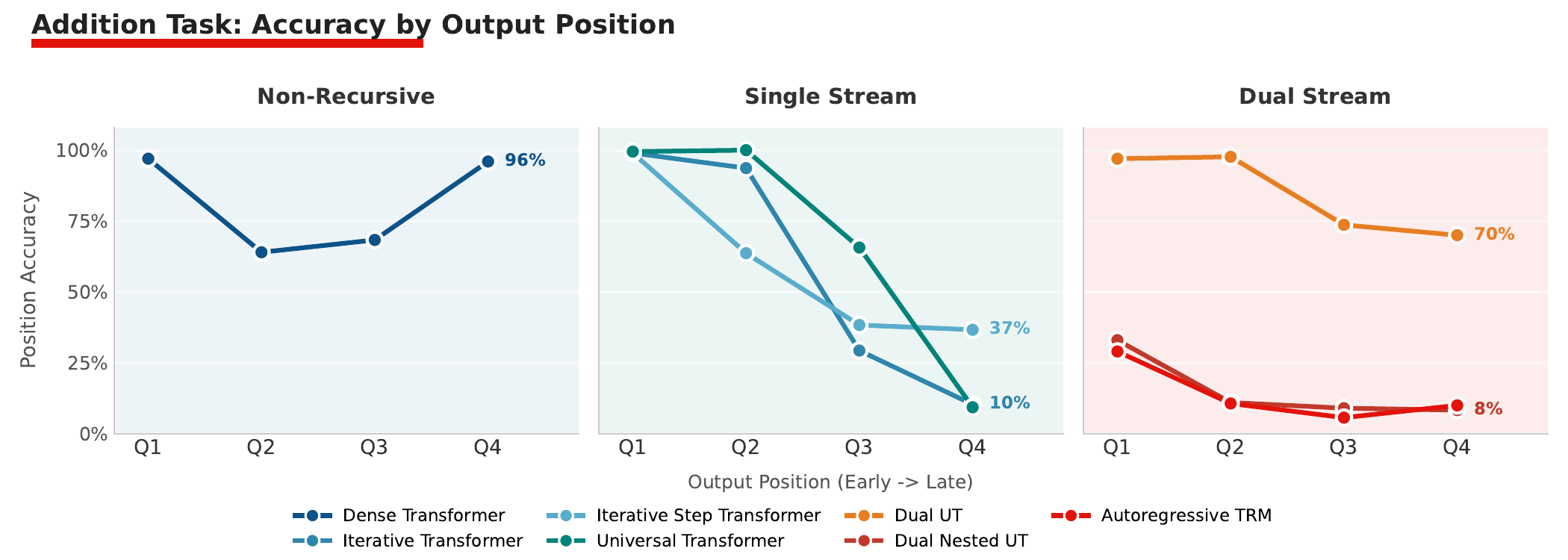}
    \caption{\textbf{Addition: accuracy stability by output position.} Character accuracy by output quartile. Dense remains high and nearly flat across positions, while several single-stream recurrent models exhibit sharp late-position collapse (Q4 $\approx$ 8--10\%). Dual UT largely avoids this collapse.}
    \label{fig:addition-position}
\end{figure*}

\paragraph{Results.}
Figure~\ref{fig:addition-position} shows that addition errors are strongly \emph{position-structured}. Dense is high and nearly flat across all quartiles, indicating stable decoding performance. In contrast, single-stream recurrent variants degrade sharply at late positions: Iterative, Iterative Step, and UT fall to roughly $\approx 10\%$, $\approx 9\%$, and $\approx 8\%$ in Q4, despite substantially higher accuracy earlier in the output. Dual UT largely avoids this late-position collapse, remaining comparatively stable across positions.

\begin{wrapfigure}{r}{0.5\textwidth}
    \centering
    \vspace{-16pt} % Adjust to align the top of the figure with the text line
    \includegraphics[width=\linewidth]{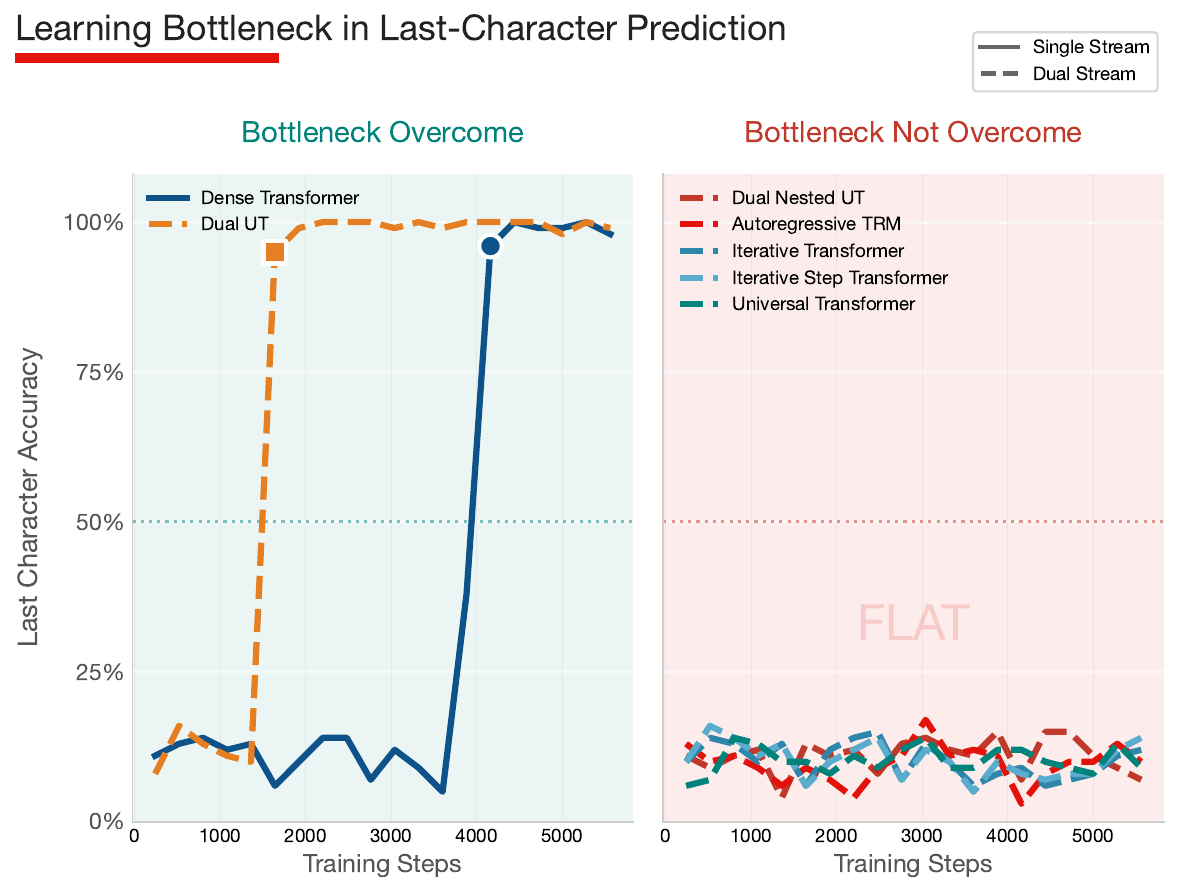}
    \caption{\textbf{Addition: training bottleneck on the final character.} Last-character accuracy over training. Only Dense and Dual UT reliably overcome the bottleneck; the remaining compute placements stay flat near chance.}
    \label{fig:last-char-bottleneck}
    \vspace{-10pt} % Adjust to pull the following text closer to the caption
\end{wrapfigure}

Figure~\ref{fig:last-char-bottleneck} shows the corresponding \emph{training-time} picture. Only Dense and Dual UT reliably break through the final-character bottleneck, with last-character accuracy rising sharply and reaching high values by the end of training. All other architectures remain essentially flat over the full training horizon: single-stream recurrent variants (Iterative, Iterative Step, UT) do not break through, and nested/terminal variants (Dual Nested UT and the Autoregressive TRM) also stay near chance.

\paragraph{Interpretation.}
Together, these plots point to addition requiring to learn and maintain a globally consistent carry-like summary across the decoding trajectory. Early output positions can often be predicted using local regularities (e.g., digitwise patterns) that do not require full consistency, so several models appear competent in Q1--Q3. The final part of the output is different: it is \emph{maximally sensitive} to upstream inconsistencies, so small representational errors that are tolerable earlier accumulate and surface as a sharp drop in Q4. We interpret this in that architectures that remain flat on last-character accuracy never acquire the core global dependency, even if they learn partial heuristics that help earlier positions. 

\begin{customblockquote}
\textbf{Takeaway.} On addition, errors concentrate late because the final dependency is least tolerant to upstream inconsistency. Whether training overcomes this bottleneck depends strongly on compute placement: Dense and flat two-stream recurrence (Dual UT) cross it, while single-stream recurrence and nested/terminal refinement typically do not.
\end{customblockquote}

\subsection{Learning dynamics}

\begin{greycustomblock}
\textbf{RQ3}. Do different placements of the same compute budget merely shift final accuracy, or do they change \emph{how} solutions are learned over training?
\end{greycustomblock}

\begin{wrapfigure}{r}{0.5\textwidth}
    \centering
    \vspace{-15pt} % Adjust to align the top of the plot with the top of the text
    \includegraphics[width=\linewidth, trim= 0cm 7cm 0cm 0cm]{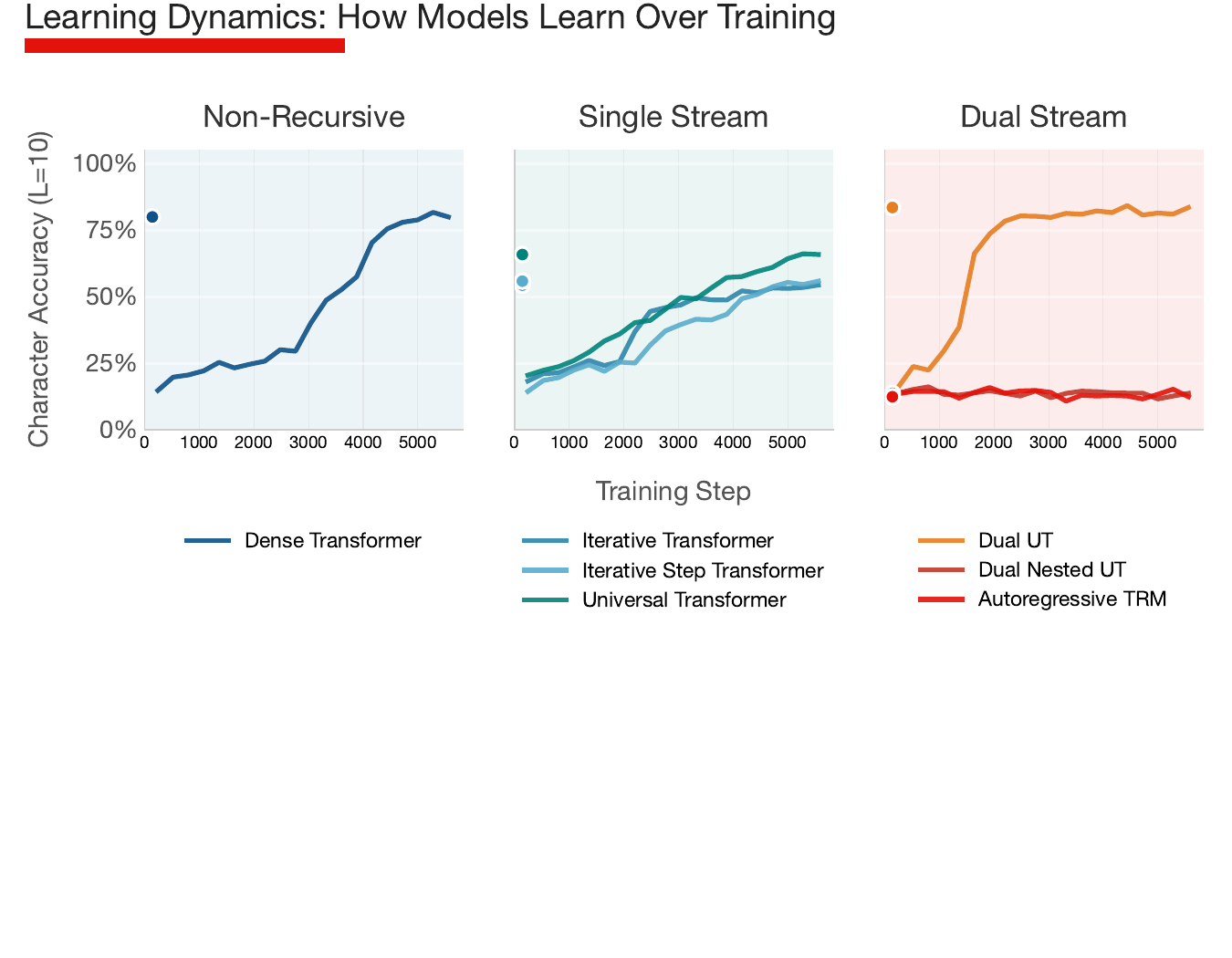}
    \caption{\textbf{Learning dynamics under compute matching (addition).} Held-out character accuracy (length $L{=}10$) over training steps. Dense shows a late, abrupt transition; single-stream tying improves gradually but plateaus; Dual UT accelerates earlier; nested/terminal variants remain low.}
    \label{fig:learning-dynamics}
    \vspace{-10pt} % Adjust to reduce white space below the caption
\end{wrapfigure}

\paragraph{Result.}
Figure~\ref{fig:learning-dynamics} shows qualitatively different learning trajectories under compute matching. Dense exhibits a long plateau followed by a late, sharp jump to high accuracy. Single-stream tied variants (Iterative, Iterative Step, UT) improve more gradually and plateau well below Dense. Dual UT accelerates earlier and reaches a higher plateau than single-stream recurrence. In contrast, nested/terminal variants (Dual Nested UT, Autoregressive TRM) remain low throughout, indicating failure to enter the high-accuracy regime. We find similar results in our further experiments, i.e. Dense models improve early positions and only later on late positions, whereas Dual UT's late positions rise more slowly and level off lower. However, we notice that the exact behavior is very sensitive to the hyperparameter and training setup used.

\paragraph{Interpretation.}
These dynamics are consistent with a shift from partial heuristics to an internal computation that resolves the global dependency in addition. Early positions admit local regularities, so models can improve without learning globally consistent carry propagation; high accuracy requires maintaining a carry-like summary that remains aligned with the readout across decoding. Untied depth may ease learning via stage-wise specialization, and flat two-stream recurrence may help by separating auxiliary computation from the solution interface. By contrast, nested refinement with terminal readout weakens credit assignment to early inner-loop steps, creating an optimization barrier that training often fails to cross.

\begin{customblockquote}
\textbf{Takeaway.} Under matched compute, compute placement changes learning dynamics: Dense and (to a lesser extent) Dual UT cross the final-dependency barrier, whereas single-stream tying often stalls and nested/terminal refinement frequently fails to learn it at all.
\end{customblockquote}

\section{Discussion}
\label{sec:discussion}
\textbf{Implications for ``latent reasoning'' claims in autoregressive models.} Our results suggest that, under a strict next-token objective with causal masking, dual stream architectures \textit{can} result in superior generalization, but the autoregressive TRM model does not exhibit such behavior. We posit that this might be attributable to either credit-assignment mechanisms intenral within different architectures or improved representation learning within a compute budget that can be exploited early on in the training. Importantly, we find that such ``dual stream'' autoregressive settings can indeed hold promise, but, at least in the small data algorithmic regime, the autoregressive TRM model does not seem to be a worthwhile research pursuit. We would be interested to see fruitful work which extends this idea further to not-so-tiny autoregressive settings or more complicated settings that require higher levels abstraction as a part of the solution space.

\paragraph{Impact Statement} This work has impact on at least several areas of machine learning and artificial intelligence, none of which we think is important to explicitly highlight.

\paragraph{Acknowledgements} We thank reviewers for their feedback and comments. This work was supported by Azure sponsorship credits granted by Microsoft’s AI for Good Research Lab.

\bibliography{paper}
\bibliographystyle{icml2026}

%%%%%%%%%%%%%%%%%%%%%%%%%%%%%%%%%%%%%%%%%%%%%%%%%%%%%%%%%%%%%%%%%%%%%%%%%%%%%%%
%%%%%%%%%%%%%%%%%%%%%%%%%%%%%%%%%%%%%%%%%%%%%%%%%%%%%%%%%%%%%%%%%%%%%%%%%%%%%%%
% APPENDIX
%%%%%%%%%%%%%%%%%%%%%%%%%%%%%%%%%%%%%%%%%%%%%%%%%%%%%%%%%%%%%%%%%%%%%%%%%%%%%%%
%%%%%%%%%%%%%%%%%%%%%%%%%%%%%%%%%%%%%%%%%%%%%%%%%%%%%%%%%%%%%%%%%%%%%%%%%%%%%%%
\newpage
\appendix
\onecolumn

\section{Extended Related Work}
\textbf{Summary}. Prior work on improving autoregressive Transformers can be understood as exploring alternative ways to allocate additional computation, but typically does so by coupling multiple design choices. Fixed-depth Transformers assign uniform compute per token and scale primarily through parameters, data, and training compute \citep{vaswani2017attention,brown2020language,hoffmann2022training}, yet both theory and empirics show that certain algorithmic or multi-step dependencies fundamentally require depth or iterative computation rather than width alone \citep{hahn2020theoretical,perez2021attention,wang2025hierarchical}, motivating approaches such as Chain-of-Thought that externalize iteration into the token stream \citep{wei2022chain}. Other work internalizes iteration by reusing parameters across steps, including Universal Transformers \citep{dehghani2018universal} and adaptive halting mechanisms such as ACT and PonderNet \citep{graves2016adaptive,banino2021pondernet}, as well as related inner-thinking or early-exit methods \citep{elbayad2019depth,xin2020deebert,chen2025inner}, though these often conflate compute allocation with halting behavior, parameter efficiency, or inference-time latency. Hierarchical and multi-timescale models further decompose latent state and update schedules, separating fast refinement from slower solution updates \citep{wang2025hierarchical,jolicoeur2025less}, continuing a broader tradition of iterative latent computation in memory-augmented and algorithmic models \citep{graves2014neural,sukhbaatar2015end,kaiser2015neural}, in contrast to rationale-based methods that render iteration explicit in generated text \citep{nye2021show,zelikman2022star}. Orthogonal approaches approximate implicit depth via fixed-point computation \citep{bai2019deep,bai2020multiscale}, allocate conditional capacity through routing in mixture-of-experts models \citep{shazeer2017outrageously,fedus2022switch}, or alter the token stream through non-monotonic or multi-pass decoding \citep{stern2019insertion,gu2019levenshtein}. Finally, some work offers refinement not via latent level processes but by iterative hypothesis generation and refinement of solutions \citep{rauba2024self}. In contrast, our work holds the autoregressive setting fixed—no routing and no token-stream modification—and isolates compute placement by directly comparing depth, recurrence, hierarchy, halting, and readout under matched block-evaluation budgets to identify where internal iteration most effectively improves generalization.

\paragraph{Fixed-depth Transformers and language modeling.}
A large body of work builds on the fixed-depth Transformer architecture \citep{vaswani2017attention}, which underlies most modern LLMs and has historically improved via scale in parameters, data, and compute \citep{devlin2019bert,radford2019language,radford2019language,brown2020language,kaplan2020scaling,hoffmann2022training}. Mechanistically, fixed depth assigns identical compute to each position, so ``easy'' and ``hard'' tokens receive the same number of block applications. This matters because constant-depth attention has formal limits for certain iterative computations: some algorithmic dependencies require depth (or a simulation of depth through additional generation steps) rather than width alone \citep{hahn2020theoretical,perez2021attention}. Empirically, even when deeper stacks help, gains can diminish on tasks that naturally require multi-step computation \citep{wang2025hierarchical}. A common workaround is to externalize iteration into the token stream—e.g., Chain-of-Thought—thereby increasing decoding length and committing intermediate steps as text \citep{wei2022chain}. In contrast, we keep the token stream fixed and study how different internal unrollings of the same decoder block change performance per unit compute.

\paragraph{Recurrent depth and adaptive computation.}
A separate line of work increases effective depth by reusing parameters across steps, turning depth into a recurrent refinement process. Universal Transformers tie layer parameters and repeatedly apply a shared block, typically augmented with step embeddings to disambiguate iterations \citep{dehghani2018universal}. Building on this, Adaptive Computation Time (ACT) introduces a learned halting unit that can stop early and often forms outputs via a weighted accumulation of intermediate states, together with a ponder-style regularizer \citep{graves2016adaptive}. PonderNet reframes halting as a probabilistic process, aiming for improved training behavior while preserving the idea of variable computation per input \citep{banino2021pondernet}. More recent ``inner thinking'' style approaches also allocate extra internal steps selectively, but typically co-vary compute allocation with additional architectural or training choices \citep{chen2025inner}. In parallel, early-exit and depth-adaptive inference methods (e.g., DeeBERT / FastBERT) stop computation once confidence is high, primarily targeting latency rather than improved generalization \citep{elbayad2019depth,xin2020deebert,liu2020fastbert}. Weight sharing is also used for parameter efficiency in settings such as ALBERT \citep{lan2019albert}; however, parameter efficiency and compute placement are distinct, since recurrence changes the computation graph even when parameter count is held fixed. Our ladder separates these effects by varying tying, step disambiguation, halting/readout, and state structure under equalized block-pass budgets.

\paragraph{Hierarchical and multi-step latent reasoning.}
Beyond flat recurrence, several approaches introduce multiple timescales or state decompositions, separating fast refinement from slower solution updates. The Hierarchical Reasoning Model (HRM) illustrates the utility of multi-rate computation for structured reasoning, suggesting that ``where iteration happens'' can matter as much as ``how much iteration happens'' \citep{wang2025hierarchical}. Tiny Recursive Models (TRM-style) similarly implement an inner loop that refines a ``reasoning'' state and an outer loop that updates a ``solution'' state, achieving strong generalization on ARC-style settings despite small parameter counts \citep{jolicoeur2025less}. These ideas connect to a longer tradition of explicitly multi-step neural computation, including multi-hop memory updates \citep{weston2014memory,sukhbaatar2015end}, iterative read/write controllers over external memory \citep{graves2014neural}, and tied-parameter procedure learning for algorithmic tasks \citep{kaiser2015neural,reed2015neural}. A key distinction is whether iteration is latent or textual: scratchpads and rationale-driven methods push multi-step computation into generated text \citep{wei2022chain,nye2021show,zelikman2022star}, whereas HRM/TRM-style models aim to keep iteration latent, enabling internal self-correction without emitting intermediate tokens. Our work sits in this latent-iteration family, but focuses on controlled comparisons: single-stream vs.\ dual-stream state, flat vs.\ hierarchical schedules, and final-state vs.\ weighted-accumulation readout, all under matched compute.

\paragraph{TRM-related work} Recently, since the publication of the TRM paper, a number of subsequent follow-up papers have appeared. These include applying TRMs for control problems \citep{jain2026tiny}, studies raising the ``block-recurrent hypothesis'' \citep{jacobs2025block}, analyses of TRMs on tabular data \citep{padayachy2026tab} or applying architectural changes to TRMs \citep{wang2026tiny, liao2026simple}.

\paragraph{Implicit depth and conditional computation.}
A different strategy is to avoid explicit unrolling and instead solve for an equilibrium. Deep Equilibrium Models compute a fixed point of a transformation and backpropagate via implicit differentiation, effectively representing infinite depth governed by solver dynamics \citep{bai2019deep}; multiscale variants extend this idea across timescales \citep{bai2020multiscale}. While these models share the theme of refinement-to-convergence, the iteration is defined by numerical solvers rather than an explicit architectural schedule, which complicates axis-by-axis attribution. Orthogonally, Mixture-of-Experts increases conditional capacity through routing rather than sequential iteration \citep{shazeer2017outrageously,lepikhin2020gshard,fedus2022switch}; we explicitly exclude routing to avoid conflating compute placement with conditional capacity. Finally, non-monotonic generation strategies refine sequences over multiple passes by inserting or editing tokens \citep{stern2019insertion,gu2019levenshtein}, and dynamic early-exit generation variants further modify decoding trajectories \citep{yoo2026adept}. These approaches demonstrate that iterative refinement can be effective, but they change the token sequence or decoding semantics relative to strict left-to-right generation. In contrast, our controlled setting holds the autoregressive family fixed—no routing, no token-stream modifications, explicit unrolling—and evaluates which placements of a fixed block-evaluation budget meaningfully improve generalization.

\FloatBarrier
\section{Experimental Framework}
\label{sec:exp_framework}

\paragraph{Compute normalization.}
We measure compute in \emph{block passes}, i.e., the number of evaluations of the decoder block executed in a single forward pass. This is the natural unit in our controlled family: it captures the dominant cost, matches the unrolling viewpoint of Section~\ref{sec:compute_placement}, and—unlike parameter count—does not confound compute with parameterization. In particular, increasing untied depth increases both the number of block evaluations and the number of distinct parameter sets, whereas increasing tied recurrence increases only the number of evaluations.

Accordingly, we fix a target budget $C$ and configure each model so that every forward pass performs exactly $C$ block evaluations. For dense (untied) depth and tied single-stream variants, one macro step consists of one block pass, and therefore $C=n$. For 2-phase variants, each macro step performs two block passes (a reasoning update followed by a solution update), giving $C=2n$. For nested variants, each macro step executes $H$ outer cycles, and within each outer cycle performs $L$ inner refinements and one solution update, so that $C=n\cdot H\cdot (L+1)$. For example, at $C=12$, a dense model uses $n=12$, a 2-phase model uses $n=6$, and a nested configuration can use $n=4$ with $(H,L)=(1,2)$.

This normalization yields the intended causal comparison: holding $C$ fixed, architectures differ only in the \emph{placement} of the $C$ block evaluations—untied depth versus tied recurrence versus within-token refinement—rather than in the number of evaluations performed.

\subsection{A controlled ladder of compute placement}
\label{subsec:ladder}

We evaluate seven architectures that form a controlled ladder. The ladder is constructed so that consecutive models differ by exactly one axis, while remaining in the same autoregressive family (tokenization, masking, objective, and block definition held fixed). Concretely, all variants are expressed as alternative unrollings of repeated applications of the same decoder-block template, differing only in (i) whether parameters are tied across applications, (ii) whether applications are step-conditioned, (iii) whether the compute budget is governed by a halting rule and whether intermediate iterates are exposed at readout, and (iv) whether the latent state is structured and updated via a flat or hierarchical schedule.

We begin with the \emph{Dense Transformer}, the standard decoder-only baseline, which composes $n$ distinct blocks sequentially and forms predictions from the final depth representation. We then isolate \emph{weight sharing} by tying all blocks, yielding the \emph{Tied Transformer} in which the same block is applied recurrently:
\[
H^{(k+1)} = f_\theta\!\big(H^{(k)}\big), \qquad k=0,\dots,K-1.
\]
Next, we isolate \emph{step disambiguation} by injecting step embeddings, giving the \emph{Tied Step-Aware Transformer}:
\[
H^{(k+1)} = f_\theta\!\big(H^{(k)} + e_k\big), \qquad k=0,\dots,K-1.
\]
From this point, we introduce \emph{adaptive compute} via \emph{Universal Transformers} with ACT \citep{graves2016adaptive}. ACT augments the recurrent unrolling with a learned halting rule and, in its standard form, exposes intermediate iterates through a weighted accumulation at readout. Because our goal is compute-equal comparison, we use ACT in a controlled regime when needed: we disable the ponder penalty and force full computation so that the realized number of block evaluations is fixed to $C$, ensuring that halting does not change the compute budget.

We then isolate the effect of \emph{state structure} by moving from a single stream to a two-stream update, producing the \emph{2-phase Universal Transformer}. Writing $X=\mathrm{Embed}(\mathbf{x})+\mathrm{Pos}(\mathbf{x})$, one macro step updates a reasoning stream $Z$ conditioned on $(X,Y,Z)$ and then updates a solution stream $Y$ conditioned on the refined reasoning state:
\[
Z' = f_\theta(X + Y + Z), \qquad Y' = f_\theta(Y + Z').
\]
This transformation places additional computation \emph{within} the autoregressive step without altering the token stream and without introducing routing. We then isolate \emph{iteration hierarchy} by nesting multiple refinements of the reasoning stream prior to each solution update (the \emph{Nested 2-phase UT}): before updating $Y$, we apply $L$ successive applications of $f_\theta$ to the reasoning stream $Z$ (each conditioned on $X$ and the current $Y$), and only then perform the solution update. Finally, we isolate the effect of \emph{readout and termination} by replacing ACT-style weighted accumulation with binary halting and final-iterate readout (the \emph{Tiny Recursive Model}, autoregressive TRM-style). In this variant, the computation halts when $\sigma(q)>0.5$, and the terminal solution iterate $Y^{(K)}$ is used to produce next-token logits, yielding a controlled instantiation of token-internal refinement beyond recurrent depth (Section~\ref{subsec:trms}).

Throughout this ladder, each architecture is obtained by adding exactly one compute-placement mechanism relative to the previous one, so that differences in performance can be causally attributed to the corresponding axis under the compute normalization of Section~\ref{sec:exp_framework}.

\subsection{Training protocol}
\label{subsec:training_protocol}

All models are trained under identical optimization and data conditions so that observed differences can be attributed to compute placement rather than to training confounds. We optimize with AdamW at learning rate $3\times 10^{-4}$, batch size 64, for a fixed number of gradient steps (1{,}000--5{,}000 across sweeps). Training sequences are generated on-the-fly from procedural generators, yielding an effectively unbounded stream of examples; this removes memorization as a viable strategy and makes generalization the primary signal through which additional compute can help.

Two implementation controls are essential for internal validity within our compute-normalized comparisons. For ACT-based models, we optionally set the ponder cost to zero and enforce full computation so that each forward pass executes the same fixed budget of $C$ block passes as the non-halting baselines, ensuring that adaptive halting does not change realized compute. Separately, when evaluating extrapolation to increased test-time compute, we optionally disable step embeddings so that increasing the number of recurrent applications does not introduce an out-of-support step signal that could confound the effect of extra iterations. 

\subsection{Tasks and evaluation}
\label{subsec:eval_tasks}

We evaluate on three character-level algorithmic tasks chosen to induce qualitatively different causal dependencies under left-to-right decoding: \textbf{Copy} (predominantly local identity), \textbf{Reverse} (long-range dependence), and \textbf{Addition} (structured multi-step dependence via carry propagation). Each task is defined by a training length $L_{\text{train}}$ and evaluated out of distribution by scaling length by fixed factors, i.e., we test at $L_{\text{eval}}=\alpha\,L_{\text{train}}$ with $\alpha\in\{1,1.5,2,3,5\}$, thereby directly probing length generalization. Decoding is greedy, and we evaluate 100 samples per condition. We report (i) sequence-level exact match, (ii) character-level accuracy, and (iii) position-wise accuracy via a quartile partition of the output positions, which localizes where errors accumulate along the generated sequence.

\subsection{Controlled variables}
\label{subsec:controlled_vars}

To satisfy the controlled-family desideratum of Section~\ref{sec:compute_placement}, we hold fixed the token stream, objective, attention semantics, and block template across all variants, so that only the \emph{placement} of block evaluations differs. Concretely, all models use the same character-level vocabulary of 14 tokens (digits 0--9, $\mid$, $+$, $=$, and newline) and are trained with next-token cross-entropy under prompt masking (loss on output tokens only). The decoder block is identical in every model: Pre-LayerNorm causal self-attention followed by a GELU MLP with $4d$ expansion and residual connections, evaluated under standard causal masking compatible with KV caching. We introduce neither routing (no Mixture-of-Experts / Mixture-of-Depths) nor token-stream modifications (no pause/think tokens and no auxiliary output streams). Under these constraints, any difference in performance arises from how the same block evaluations are scheduled and exposed, enabling causal attribution in Section~\ref{sec:evaluation} to specific compute-placement mechanisms.

\end{document}